%% file: emnlp2020.tex
\pdfoutput=1
\documentclass[11pt,a4paper]{article}
\usepackage[hyperref]{emnlp2020}
\usepackage{times}
\usepackage{latexsym}
 
\usepackage{microtype}

\usepackage{multirow}
\usepackage{amsmath}
\usepackage{amsthm}
\usepackage[pdftex]{graphicx}
\graphicspath{{figs/}}
\usepackage{amssymb}
\usepackage{subcaption}

\usepackage{arydshln}
\usepackage[T1]{fontenc}

\aclfinalcopy 

\title{End-to-End Synthetic Data Generation for Domain Adaptation of Question Answering Systems}

\author{Siamak Shakeri$^{\dagger*}$, Cicero Nogueira dos Santos\thanks{*equal contribution. $\dagger$ Siamak Shakeri is currently with Google. The work was done when he was at AWS AI.} , Henry Zhu, Patrick Ng, \\
\textbf{Feng Nan, Zhiguo Wang, Ramesh Nallapati, Bing Xiang}  \\
  AWS AI \\
  New York City, NY \\
  \texttt{\{cignog,henghui,patricng,nanfen\}@amazon.com} \\
  \texttt{\{zhiguow,rnallapa,bxiang\}@amazon.com} \\
  \texttt{siamaks@google.com} \\
}

\date{}

\begin{document}
\maketitle
\begin{abstract}
We propose an end-to-end approach for synthetic QA data generation.
Our model comprises a single transformer-based encoder-decoder network that is trained end-to-end to generate both answers and questions.
In a nutshell,
we feed a passage to the encoder and ask the decoder to generate a question and an answer token-by-token.
The likelihood produced in the generation process is used as a filtering score, which avoids the need for a separate filtering model.
Our generator is trained by fine-tuning a pretrained LM using maximum likelihood estimation. The experimental results indicate significant improvements in the domain adaptation of QA models outperforming current state-of-the-art methods.
\end{abstract}

\input{introduction.tex}
\input{method.tex}
\input{related_work.tex}

\input{experiments.tex}
\input{conclusion.tex}

\bibliography{emnlp2020}
\bibliographystyle{acl_natbib}
\newpage
\input{appendix.tex}

\end{document}

%% file: introduction.tex
\section{Introduction}
Improving question answering (QA) systems through automatically generated synthetic data is a long standing research goal \citep{mitkov2003computer,rus2010first}.
Although many past works have proposed different strategies for question generation, they have limited or no success in improving the downstream QA task  \cite{du-etal-2017-learning,sun-etal-2018-answer,song-etal-2018-leveraging,klein2019learning,wangneural,ma2020improving,Chen2020Reinforcement,tuan2020capturing}.

Some recent approaches for synthetic QA data generation based on large pretrained language models (LM) have started to demonstrate success in improving the downstream Reading Comprehension (RC) task with automatically generated data \cite{alberti2019qgen,puri2020nvidia}.
However, these approaches typically consist of multi-stage systems that use three modules: 
span/answer detector, question generator and question filtering.
Given an input passage, 
the \emph{span detector} is responsible for extracting spans that will serve as answers for which questions will be generated. This module normally combines a pretrained QA model with handcrafted heuristics.
The \emph{question generator} is a large LM fine-tuned for the task of conditional generation of questions given passage and answer.
The \emph{question filtering} comprises another RC model that is used to score and filter the generated QA pairs.
Each module of this synthetic data generation pipeline is trained/tuned separately and errors from one stage can propagate to the next stages.
Additionally,
each module is expensive to be computed because all use large transformer networks \cite{vaswani17_transformer}.

In this work,
we propose an end-to-end approach for synthetic QA data generation.
Our model comprises a single transformer-based encoder-decoder network that is trained end-to-end to generate both the answer and the question.
In a nutshell,
we feed a passage to the encoder and ask the decoder to generate the question and the answer token-by-token.
The likelihood produced in the generation process is used as a filtering score, which avoids the need of a separate filtering model.
Our generator is trained by fine-tuning a pretrained LM using maximum likelihood estimation (MLE).
We use BART \cite{lewis2019bart} as the pretrained LM in our experiments.

We perform experiments with three different variations of our synthetic QA data generator: (1) \emph{AQGen}, which generates first the answer then the question; (2) \emph{QAGen}, which generates first the question then the answer; (3) \emph{QAGen Two-step (2S)}, which generates first the question, concatenates it to the passage, then generates the answer in a second pass through the same encoder-decoder.

We focus our empirical evaluation on the task of data augmentation for domain adaptation of reading comprehension (RC) models trained on SQuAD 1.1 dataset.
We assess the effectiveness of our QA data generators for domain adaptation of four different target domain datasets: Natural Questions (NQ), BioASQ, NewsQA and DuoRC. 
We compare our results with recent work on domain adaptation for QA as well as with a three-stage synthetic data generator.
QAGen performs better than AQGen and the baselines for all datasets, while QAGen2S provides the best results overall because it allows bidirectional attention between passage and question.
For NQ dataset, QAGen2S improves the SQuAD baseline by more than 8 points in EM and more than 7 points in F1. For NewsQA and BioASQ the gains in EM are also above 4 points.
Additionally, we also demonstrate that synthetically generated data by QAGen2S can improve the in-domain performance of both small and large RC models, leading to F1/EM improvements of 1/0.5 and 3.1/2.2 on \texttt{RoBERTa-large} and \texttt{bert-base-uncased} trained RC models on SQuAD dev.

The main contributions of this work can be summarized as follows: 
(1) we propose the first effective end-to-end approach for synthetic QA data generation;
(2) our approach solves an important issue in previous methods for QA data generation: the detection of good spans. We show that span detection can be effectively solved as a generation task, just like question generation;
(3) as it uses a single end-to-end model, our data generation pipeline is simpler, faster and more efficient;
(4) we perform comprehensive experiments that demonstrate the effectiveness of our proposed approach for domain adaptation of QA systems.

%% file: method.tex
\section{End-to-End Model for Question and Answer Generation and Filtering}
We model the problem of synthetic QA data generation as a conditional language modeling task.
More specifically,
we use an encoder-decoder (enc-dec) conditional LM as described in what follows.
\subsection{Enc-Dec Conditional Language Models}
Language modeling consists of learning the probability distribution $p(x)$ over variable-length token sequences $x=(x_1,x_2,...,x_{|x|})$,
where the tokens come from a fixed size vocabulary $V$.
The training of LMs typically involves solving the task of predicting the next token based on past tokens.
The distribution $p(x)$ can be represented by the conditional probability of the next token given the previous ones \cite{bengio:2003}:
\begin{equation}
    p(x)= \prod^{|x|}_{i=1}{p(x_i|x_{<i})}
\end{equation}
In the case of conditional LMs, the generation is conditioned on an additional context $c$:
\begin{equation}\label{eq:conditional_lm}
    p(x|c)= \prod^{|x|}_{i=1}{p(x_i|x_{<i},c)}
\end{equation}

Transformer-based encoder-decoder conditional LMs \cite{lewis2019bart,raffel2019t5} use bidirectional self-attention in the encoding step to create vector representations of the tokens in the context $c$.
The decoding step generates the tokens of the sequence $x$ in an auto-regressive manner,
while performing self-attention on previously generated tokens of $x$ and all the representations output by the encoder for $c$.

\subsection{Question-Answer Generation}
In the case of end-to-end synthetic data generation for QA, 
we need to model the joint conditional distribution $p(a,q|c)$, 
where the input context $c$ is a passage,  
$q$ is a question and $a$ is the correct answer, which is a span in $c$.
Our approach to model $p(a,q|c)$ involves fine-tuning a pretrained Enc-Dec conditional LM using a training set \mbox{$D=\{(c^1,q^1,a^1), (c^2,q^2,a^2), ... , (c^{|D|},q^{|D|},a^{|D|})\}$}.
We train the Enc-Dec with parameters $\theta$ through maximum likelihood estimation (MLE) by minimizing the negative log-likelihood over $D$: 
\begin{equation}
    \mathcal{L}(D)= -\sum^{|D|}_{i=1} {\log p_{\theta}(a^i,q^i|c^i)}
\label{eq:MLE}
\end{equation}

\begin{figure}[htbp]
 \centering
  \includegraphics[width=0.33\textwidth,]{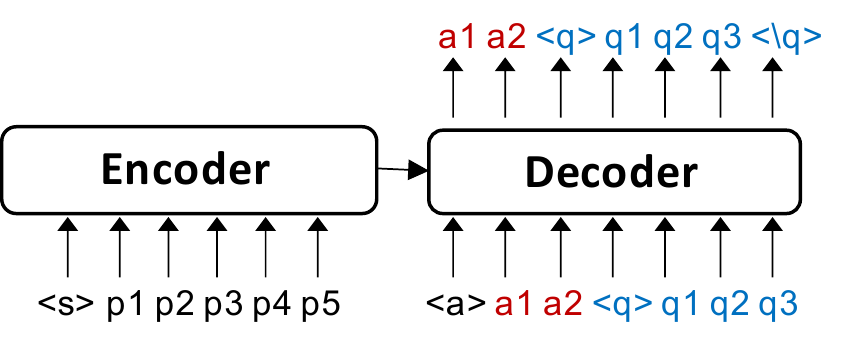}
 \caption{\small AQGen Model: given an input passage the model generates an answer followed by a question.}
 \label{figs:AQGen}
\end{figure}

We can have different variations of the generator depending on how we place the items in the output sequence: answer-question or question-answer.
This difference in the ordering is crucial because it defines which part is conditioned on the other.
Based on this observation, we propose three variations of our generative model: 

\textbf{AQGen}: this model generates answer and question jointly given the input context: $(q,a) \sim p(a , q | c)$. During sampling, the answer tokens are generated, which are followed by question tokens. This makes the generation of the question conditioned on both input context (through attention on the encoder) and answer (through self-attention in the decoder). Fig. \ref{figs:AQGen} depicts this model.

\begin{figure}[htbp]
 \centering
  \includegraphics[width=0.33\textwidth]{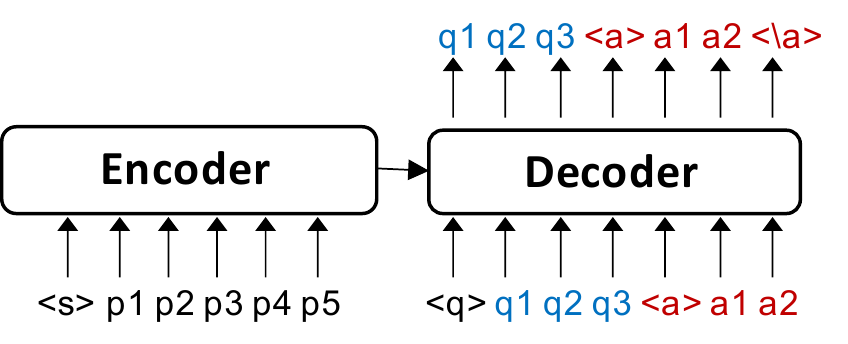}
 \caption{\small  QAGen Model: given an input passage the model generates a question followed by an answer.}
 \label{figs:QAGen}
\end{figure}

\textbf{QAGen}: this model generates question and answer jointly given the input passage: $(q,a) \sim p(a , q | c)$. During sampling, the question tokens are generated, which are followed by answer tokens. This makes the generation of the answer conditioned on both input context (through attention on the encoder) and question (through self-attention in the decoder). Fig. \ref{figs:QAGen} depicts this model. 

\begin{figure}[htbp]
 \centering
  \includegraphics[width=0.33\textwidth]{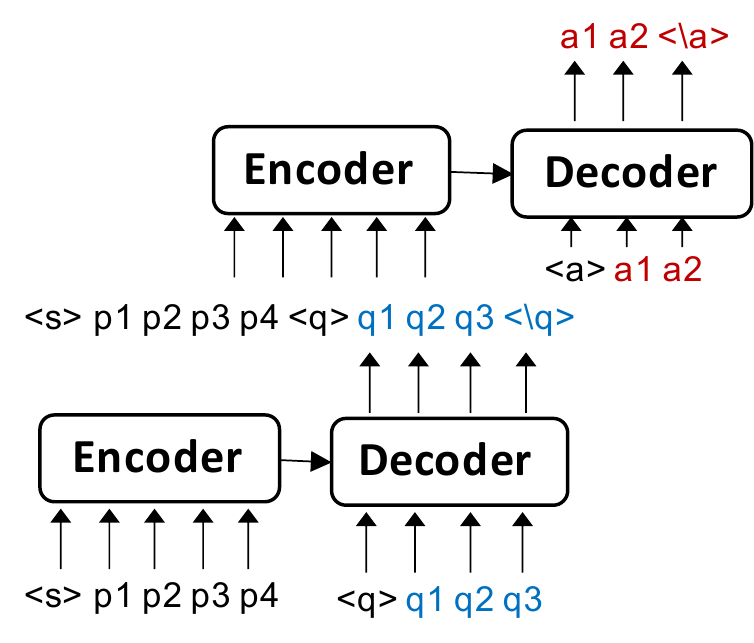}
 \caption{\small QAGen Two-Step: given an input passage the model first generates a question (Step 1). Next, the question is concatenated with the passage and both are given to the encoder-decoder that generates the answer (Step 2).}
 \label{figs:twostep}
\end{figure}

\textbf{QAGen Two-Step (2S)}: this model performs question generation and answer generation in two separate passes over the Enc-Dec LM.
First,
the question is generated given the input context $q \sim p(q | c) $,
(Step 1).
Next,
the question is concatenated with the input context and the resulting sequence is given as input to the Enc-Dec,
which finally generates the answer $a \sim p(a | q, c)$, (Step 2).
QAGen 2S sampling approach is illustrates in Fig. \ref{figs:twostep}.
This model uses a single Enc-Dec LM that is trained with samples of both $p(q | c) $ and $p(a | q, c)$. We use control codes \emph{$<$q$>$} and \emph{$<$a$>$} to inform the decoder whether to generate a question or an answer, respectively.


    


 \begin{table*}[ht]
\centering
\scriptsize
\begin{tabular}{ll}
 \hline
\multicolumn{2}{l}{}
\begin{minipage}{15 cm}
\textit{\textbf{PubMed}}\\ Lymph node status has major prognostic importance in colorectal cancer and greater precision in the diagnosis of lymph node metastases should provide better prognostic and therapeutic guidance. \textbf{Keratin 20} (K20) gene expression has been used as a marker of lymph node metastases, but the evidence for this remains circumstantial. This study has therefore sought to determine K20 specificity and to correlate K20 expression with mutant K-RAS expression, in order to provide direct evidence that K20 expression in lymph nodes of colorectal cancer patients genuinely reflects metastatic disease. Specificity of K20 expression was established against a range of tissue types and 289 lymph nodes from 41 non-cancer control patients. K20 expression was restricted to \textbf{gastrointestinal epithelia} and was only present in one of the 289 control lymph nodes, giving a calculated specificity of 97.6 $\%$ (95$\%$ confidence limits: \textbf{87.1-99.9$\%$})...

  \end{minipage} \hfill
  \\
 \hdashline
 Q: \textit{What is K20 expression found to be restricted to?} & A: \textit{gastrointestinal epithelia}\\ 
 Q: \textit{What was the 95$\%$ confidence range of the mutation analysis?} & A:  \textit{87.1-99.9$\%$} \\ 
 Q: \textit{What is the name of the gene that can be used as a marker of metastatic disease?} & A: \textit{Keratin 20} \\
 \hline
 \multicolumn{2}{l}{}
\begin{minipage}{15 cm}
\textit{\textbf{CNNDM}}\\ 
By. Emily Allen. PUBLISHED:. 06:27 EST, 12 June 2012. |. UPDATED:. 09:35 EST, 12 June 2012. Teachers have apologised to parents after a group of \textbf{primary school} children were forced to stay in the canteen until they had finished all the food on their plates. Parents of children attending Kaizen Primary School in Plaistow, East London, were left fuming after a group of pupils, some as young as \textbf{five}, were told they had to clear their plates before being allowed out into the playground. Even though years ago parents would not have batted an eyelid and would have welcomed schools encouraging their children to eat, dozens of parents complained, saying that children should 'not be forced to eat' by teachers. Upset: Parents of children at Kaizen Primary School in Plaistow, East London, said pupils were told they had to clear their plates (file picture) Candeece Kenlock said her five-year-old son \textbf{Kehyan} was 'so scared' of being forced to eat everything on his plate he didn't want to go to school anymore....
  \end{minipage} \hfill
  \\
 \hdashline
Q: \textit{what is the name of a five year old boy whose parents said}  & A: \textit{Kehyan}\\ 
 \textit{he was 'so scared' he didn't want to go to school?} \\
Q: \textit{What type of school were children forced to stay} & A:  \textit{primary school} \\ 
 \textit{in the canteen to finish their meals?} \\
Q: \textit{How old were the children who were forced to stay in}  & A: \textit{five} \\
 \textit{the canteen until they had finished their food?} \\
 \hline
 
  \multicolumn{2}{l}{}
\begin{minipage}{15 cm}
\textit{\textbf{IMDB}}\\ 
 \textbf{Clark Russell}, a prominent writer, concludes that he will visit the south in the capacity of a farm hand and thus secure atmosphere for a new story. He learns that laborers are needed on a certain farm and as he journeys into the country he rescues a young woman whose horse is running away. When Clark applies for work he is treated lightly by \textbf{Bud}, the foreman, until the owner of the farm arrives with his daughter, Anna, who recognizes her hero of the afternoon. A few days later at the dinner table Clark defends Polly, a maid, when she is annoyed by Bud and after the hands departed for the fields the two men settle their score in a fight, the bully receiving a severe lesson. Polly overhears Bud declaring that he will be revenged but she is unable to warn Clark. Later in the day \textbf{the bully} tries to force Clark into the hopper of the threshing machine but Anna sees the struggle from a distance and stops the engine...
  \end{minipage} \hfill
  \\
 \hdashline
Q: \textit{What is the name of the foreman at the farm?}  & A: \textit{Bud}\\ 
Q: \textit{Who saves Anna?} & A:  \textit{Clark Russell} \\ 
Q: \textit{Who tries to force Clark into a hopper of the threshing machine?}  & A: \textit{the bully} \\
 \hline
 
 \multicolumn{2}{l}{}
\begin{minipage}{15 cm}
\textit{\textbf{Natural Questions}}
\\
<Table> <Tr> <Th colspan="2"> Tampa Bay Lightning </Th> </Tr> <Tr> <Td colspan="2"> 2018 -- 19 Tampa Bay Lightning season </Td> </Tr> <Tr> <Td colspan="2"> </Td> </Tr> <Tr> <Th> Conference </Th> <Td> Eastern </Td> </Tr> <Tr> <Th> Division </Th> <Td> Atlantic </Td> </Tr> <Tr> <Th> Founded </Th> <Td> \textbf{1992} </Td> </Tr> <Tr> <Th> History </Th> <Td> Tampa Bay Lightning 1992 -- present </Td> </Tr> <Tr> <Th> Home arena </Th> <Td> Amalie Arena </Td> </Tr> <Tr> <Th> City </Th> <Td> Tampa , Florida </Td> </Tr> <Tr> <Td colspan="2"> </Td> </Tr> <Tr> <Th> Colors </Th> <Td> Tampa Bay blue , white </Td> </Tr> <Tr> <Th> Media </Th> <Td> Fox Sports Sun 970 AM </Td> </Tr> <Tr> <Th> Owner ( s ) </Th> <Td> Tampa Bay Sports and Entertainment ( Jeffrey Vinik , chairman ) </Td> </Tr> <Tr> <Th> General manager </Th> <Td> \textbf{Steve Yzerman} </Td> </Tr> <Tr> <Th> Head coach </Th> <Td> \textbf{Jon Cooper} </Td> </Tr> <Tr> <Th> Captain </Th> <Td> Steven Stamkos </Td> </Tr> <Tr> <Th> Minor league affiliates </Th> <Td> Syracuse Crunch ( AHL ) Orlando Solar Bears ( ECHL ) </Td> </Tr> <Tr> <Th> Stanley Cups </Th> <Td> 1 ( 2003 -- 04 ) </Td> </Tr> <Tr> <Th> Conference championships </Th> <Td> 2 ( 2003 -- 04 , 2014 -- 15 ) </Td> </Tr> <Tr> <Th> Presidents ' Trophy </Th> <Td> 0 </Td> </Tr> <Tr> <Th> Division championships </Th> <Td> 3 ( 2002 -- 03 , 2003 -- 04 , 2017 -- 18 ) </Td> </Tr> <Tr> <Th> Official website </Th> <Td> www.nhl.com/lightning </Td> </Tr> </Table>
\\
  \end{minipage} \hfill \\
   \hdashline
 Q: \textit{What year was the Tampa Bay Lightning established??} & A: \textit{1992}\\ 
 Q: \textit{Who is the head coach of the Tampa Bay Lightning?} & A:  \textit{Jon Cooper} \\ 
 Q: \textit{Who is the Tampa Bay Lightning general manager?} & A:  \textit{Steve Yzerman}\\ 
\end{tabular}
\caption{\small Samples of generated question-answer pairs using QAGen2S model for four target domains. The generated answers are shown in \textbf{bold}. The paragraphs are truncated from their original sizes due to space limitations.}
\label{tab:samples}
\end{table*}
\subsection{Decoding}\label{sec:decoding}
A natural choice for decoding with conditional neural LMs is beam search.
However, our preliminary experiments with beam search showed a lack of diversity and a high repetition of generated question-answer pairs.
Generating diverse question-answer pairs is crucial to the performance of downstream RC models. Particularly, diversity of answer spans ensures that various parts of the passage are used, and different question types are generated. 
We use a variant of nucleus sampling \cite{nucleus}, where we pick top k tokens, and within top k, we pick tokens that comprise top 95$\%$ probability mass. We set k to 20 in our experiments. We refer to this setting as \textit{Topk+Nucleus}. This decoding was used in QAGen, AQGen, and question sampling step in QAGen2S. The answer generation of QAGen2S was performed by greedy decoding. We discard generated $(q, a)$ pairs whose answers do not occur in the input passage, as non-extractive QA is outside the scope of this work. We observed between 10$\%$ to 15$\%$ of samples being dropped because of this issue.

\subsection{Filtering}\label{sec:filtering}
Recent work have used the \textit{round-trip filtering} method \cite{alberti2019qgen,puri2020nvidia} to prune the synthetic QA set and improve data quality.
This method consists of two steps: (1) using an RC model to provide answers to the generated questions;
(2) dropping the QA pairs for which the answer of the RC model does not match the span detected answer. While round-trip filtering has shown to be effective, it is not the most efficient approach because it involves the application of an RC system over the whole set of generated data.
Additionally, there might exist cases that are difficult for the filtering model, but in fact are of high quality.

We propose using the likelihood of the generated question-answers as a measure to perform filtering and address the efficiency issue,
as it avoids the use of an RC model for filtering. We argue that such a likelihood score, albeit noisy, is an indicator of whether a generated question-answer is high quality for training a downstream RC model. We refer to this approach as \textit{LM filtering}.
Essentially, given an input passage, we sample $n$ different QA pairs, rank them according to decreasing order of LM score and pick the top $m$ samples. 
This is similar to the \emph{sample-and-rerank} approach suggested by \citet{nucleus} and \citet{adiwardana2020meena}.
Formally, for QAGen and QAGen2S, we use the score:
\begin{equation*}
     \text{LM score}= \sum^{N_{a}}_{i=1} {\log p(a^i|c,q)}
     \label{eq:lm1}
\end{equation*}
And for AQGen :
\begin{equation*}
    \text{LM score}= \sum^{N_{a}}_{i=1} {\log p(a^i|c)} + \sum^{N_{q}}_{i=1} {\log p(q^i|c,a)}
    \label{eq:lm2}
\end{equation*}
Where $N_q$ and $N_a$ indicate the lengths of generated question and answer, respectively.
We use answer-only scores for QAGen and QAGen2S because question quality would have a dominant effect on LM scores since questions are usually longer than answers.
Additionally, using answer-only scores when conditioned on the generated question is more suitable for the RC tasks because it better mimics the score of a downstream RC model, which is answer centric. With AQGen, we use both answer and question LM scores, as answer generation is not conditioned on the question.
We use likelihood summation instead of averaging because experiments showed that the former works slightly better. Further details included in Appendix \ref{ap:sec:lm_score_pooling}.
We speculate this is due to average pooling encouraging longer question-answers, which could be of lower quality than shorter question-answer pairs.

%% file: related_work.tex
\section{Related Work}
Question generation (QG) has been extensively studied from the early heuristic-based methods \citep{mitkov2003computer,rus2010first} to the recent neural-base approaches. However, most work \citep{du-etal-2017-learning,sun-etal-2018-answer,zhao-etal-2018-paragraph,kumar-etal-2019-putting,wangneural,ma2020improving,tuan2020capturing,Chen2020Reinforcement} only takes QG as a stand-alone task, and evaluates the quality of generated questions with either automatic metrics such as BLEU, or human evaluation. \citet{tang2017question}, \citet{duan-etal-2017-question} and \citet{sachan-xing-2018-self} verified that generated questions can improve the downstream answer sentence selection tasks. \citet{song-etal-2018-leveraging} and \citet{klein2019learning} leveraged QG to augment the training set for machine reading comprehend tasks. However, they only got improvement when only a small amount of human labeled data is available. Recently, with the help of large pre-trained language models, \citet{alberti2019qgen} and \citet{puri2020nvidia} have been able to improve the performance of RC models using generated questions. However, they need two extra BERT models to identify high-quality answer spans, and filter out low-quality question-answer pairs.  \citet{vaeQA} follow a similar approach while using InfoMax Hierarchical Conditional VAEs. \citet{nishida2019uda} showed improvements  by fine-tuning the language model on the target domains.



%% file: experiments.tex
\section{Experimental Setup and Results}
\subsection{Datasets}
We used \textbf{SQuAD 1.1} dataset \cite{squad11} to train the generative models as well as in-domain supervised data for the downstream RC task in this work. We used the default train and dev splits, which contain 87,599 and 10,570 $(q,a)$ pairs, respectively.

Similar to \cite{nishida2019uda}, we selected the following four datasets as target domains:\\
\textbf{Natural Questions} \cite{nq_dataset}, which consist of Google search questions and the annotated answers from Wikipedia. 
We used MRQA Shared Task \cite{fisch2019mrqa} preprocessed training and dev sets, which consist of 104,071 and 12,836 $(q,a)$ pairs, respectively. The training set passages were used as the unlabeled target domain corpus, while the evaluations were performed on the  dev set.\\
\textbf{NewsQA} \cite{newsqa_dataset}, which consists of question and answer pairs from CNN news articles. We used the dev set from the MRQA Shared Task, which removes unanswerable questions and those without annotator agreement. We prefer this version as we focus only on the generation of answerable questions. The dev set consists of 4,212 $(q,a)$  pairs. Passages from CNN/Daily Mail corpus of \citet{newsqa_dataset} are used as unlabeled target domain corpus.\\
\textbf{BioASQ} \cite{bioasq}: 
we employed MRQA shared task version of BioASQ, which consists of a dev set with 1,504 samples. We collected PubMed abstracts to use as target domain unlabeled passages. \\
\textbf{DuoRC} \cite{DuoRC} contains question-answer pairs from movie plots which are extracted from both Wikipedia and IMDB.
ParaphraseRC task of DuoRC dataset was used in our evaluations, consisting of 13,111 pairs. We crawled IMDB movie plots to use as the unlabeled target domain corpus.
\subsection{Experimental Setup}
We used Pytorch \cite{pytorch} and Transformers \cite{HuggingFacesTS} to develop the models and perform experiments.
Generative models are trained on SQuAD 1.1 for 5 epochs, and the best model is selected based on the cross entropy loss on the SQuAD dev set. AdamW \cite{AdamW} optimizer with learning rate of $3\times 10^{-5}$ is employed.

For RC model training, we use \texttt{bert-base-uncased} model \cite{bert}. AdamW optimizer is used with learning rate of $3\times 10^{-5}$ and batch size 24 for 2 epochs without linear warmup. We set maximum sequence length 384 with document stride 128. SQuAD 1.1 dev set is used to select the best model during training.
As a baseline for QA data generation, we implemented a three-stage pipeline similar to the state-of-the-art approach of \citet{puri2020nvidia}. We call this baseline \emph{\textbf{QGen}}, which generates a question given a passage and extracted span, $q \sim p(q | a, c)$. The span detection module consists of \texttt{bert-base-uncased} fine-tuned on SQuAD 1.1 passage and spans, where the start and end classification heads are trained to perform span detection.
For QGen, we experimented with sampling top 5 spans and generating two questions per each, as suggested by \cite{puri2020nvidia}, as well as sampling top 10 spans and generating one question per each. Our results showed the latter outperforming the former. Henceforth, we used this configuration in our evaluations.

We trained QGen models on both BART-Large and GPT2-Medium \cite{radford2019gpt2}, which have an equivalent number of parameters, 406M (BART) vs 350M (GPT2), and evaluated BLEU score of the generated question w.r.t. the ground truth question on the SQuAD dev set. BART and GPT2 achieved 21.29 and 18.31 BLEU, respectively. We believe the bi-directional encoding in BART is superior to uni-directional encoding in GPT2. Hence, we used BART for the rest of the experiments.
\subsection{Synthetic Data Generation}
For each of the unlabeled target domain corpora, we randomly selected 100,000 passages to perform synthetic data generation. Passages shorter than 100 tokens were discarded. Selected ones were truncated to maximum length of 550 tokens. We removed the passages that existed in the dev sets.

Question-answer generation with AQGen, QAGen, and QAGen2S is performed using Topk+Nucleus, as discussed in Sec. \ref{sec:decoding}. For each passage, 10 samples are generated. Unless otherwise mentioned, LM filtering is applied by sorting the 10 samples of each passage according to LM scores as detailed in Sec. \ref{sec:filtering}, and the top 5 samples are selected.
The number of synthetically generated pairs is between 860k to 890k without filtering and 480k to 500k after LM filtering.
Tab. \ref{tab:samples} shows generated question-answer pairs from four target domain (see Appendix for more examples). We can observe that the generative model is able to generate question answer pairs even from raw HTML input that corresponds to a table. The rendered table can be seen in Tab. \ref{tab:table_samples} (Appendix \ref{ap:sec:qa_from_table}). Considering the fact that the training data of the generative model does not include any HTML input, this further demonstrates the robustness and efficacy of our proposed approach.
\begin{table*}[ht]
\centering
\footnotesize
\scalebox{0.92}{
\begin{tabular}{lc|cccccccc}
\hline
\multirow{2}{*}{\textbf{Model}} & \textbf{fine-tune} & \multicolumn{2}{c}{\textbf{NQ}} & \multicolumn{2}{c}{\textbf{NewsQA}} & \multicolumn{2}{c}{\textbf{BioASQ}} &
\multicolumn{2}{c}{\textbf{DuoRC}} \\ 
& \textbf{Data} & EM & F1 & EM & F1 & EM & F1 & EM & F1\\
\hline
SQuAD 1.1 \citet{nishida2019uda} & SQuAD  & 44.4 & 57.5 & 35.2 & 50.7 & 41.1 & 53.6 & 24.5 & 33.0\\
UDA \citet{nishida2019uda} & SQuAD & 43.8 & 56.7  & 35.9 & 51.4  & 45.4 & 57.8 &  25.5 & 34.1\\ 
\hline
SQuAD 1.1 \citet{vaeQA} & SQuAD  & 42.77 & 57.29 &-- & -- & -- & -- & -- & --\\
UDA \citet{vaeQA} &  SQuAD+Synthetic & 48.44 & 62.69  & -- & --  & -- & -- &  -- & --\\ 
\hline
Our SQuAD 1.1 & SQuAD & 44.66 & 58.94
   & 39.51 &56.36 & 44.35 & 56.06 & 28.85 &34.92\\ \hdashline
\multirow{2}{*}{QGen + round-trip filtering}  & Synthetic& 48.04 & 61.28 & 39.03 & 54.37 & 35.31 & 46.80 & 28.74 & 34.10 \\
                       & + SQuAD &  49.02 & 62.61 & 40.79 & 56.79 & 39.43 & 50.42 & 29.39 & 34.80\\ 
\hline
\multirow{2}{*}{AQGen (ours) + LM filtering}  & Synthetic   & 47.80 & 61.29 & 38.55 & 55.42 & 39.49 &52.11 & 27.09	& 33.47  \\ 
                       & + SQuAD & 49.04 & 62.56 & 39.62 & 56.88 & 42.89 & 54.90 & 27.88 & 34.40  \\ \hdashline
\multirow{2}{*}{QAGen (ours) + LM filtering}  & Synthetic   & 49.81 &63.36 & 43.09 & 57.9 & 42.49 & 51.95& 29.46 &35.25 \\ 
                       & + SQuAD &  50.01 & 63.10 & 44.06 &	59.20 & 45.74 &55.06 & 29.91	& 35.82 \\ \hdashline
\multirow{2}{*}{QAGen2S (ours) + LM filtering}  & Synthetic   
 & 52.64 &65.56& 43.99 & 59.95 & 46.74 & 57.76 & 29.91 & 35.81 \\
& + SQuAD  & 52.03 & 65.70 & 43.57 & 59.8 & \textbf{48.40} & \textbf{58.33} & \textbf{30.06}& \textbf{36.05} \\
\hline
\multirow{2}{*}{QAGen2S (ours) + round-trip}  & Synthetic &  \textbf{53.11} & \textbf{66.45}  &  \textbf{45.04} & 60.79 & 45.01 & 57.01 & 29.47 & 35.32\\
                       & + SQuAD & 51.91 & 65.62 & 44.78&  \textbf{60.92} & 46.14 & 57.96 & 30.01 & 35.83 \\ 
\hline
Supervised target domain  & Target  & 66.50	& 78.55 & 51.09	& 66.67 & -- & -- & 27.35 & 33.28
\\
\hline
\end{tabular}
}
\caption{{\small Domain adaptation results for different methods. \textbf{Bold} cells indicate the best performing model on each of the target domain dev sets, excluding supervised target domain training results.
}}\label{tab:uda}
\end{table*}

\begin{table*}[ht]
\centering
\footnotesize
\scalebox{0.92}{
\begin{tabular}{lc|cccccccccc}
\hline
\multirow{2}{*}{\textbf{Target Domain Corpus}} & \textbf{fine-tune} & \multicolumn{2}{c}{\textbf{NQ}} & \multicolumn{2}{c}{\textbf{NewsQA}} & \multicolumn{2}{c}{\textbf{BioASQ}} &
\multicolumn{2}{c}{\textbf{DuoRC}} &
\multicolumn{2}{c}{\textbf{SQuAD}} \\
& \textbf{Data} & EM & F1 & EM & F1 & EM & F1 & EM & F1 & EM & F1 \\ \hline
SQuAD 1.1 & SQuAD & 44.66 & 58.94
   & 39.51 &56.36 & 44.35 & 56.06 & 28.85 &34.92 & 80.78 & 88.20 \\ 
\hline
\multirow{2}{*}{Natural Questions}  & Synthetic   &\underline{\textit{52.64}} & 65.56 & 40.48&55.40& 42.69&52.56 & 27.88&33.39 & 79.95 & 86.89 \\
                       & + SQuAD & 52.03 & \underline{\textit{65.70}} & 40.55&56.37&44.15&55.87 & 30.04 & \underline{\textit{36.14}} & 83.05 & 89.91\\ \hdashline
\multirow{2}{*}{CNN/DM}  & Synthetic & 47.05 & 60.27& \underline{\textit{43.99}}	& \underline{\textit{59.95}}&45.28 & 55.25& 27.02 & 33.22 & 76.81 & 84.62\\
                       & + SQuAD &45.92	& 60.24 & 43.56 & 59.8 & 44.88&57.06&27.62 & 34 & 82.29 & 89.32 \\\hdashline
\multirow{2}{*}{PubMed}  & Synthetic   
    & 44.48&57.98 & 39.27 & 54.88 & 46.74 & 57.76 & 26.21 & 32.03 & 78.65 & 85.82 \\
 & + SQuAD & 48.08&61.73 
   & 41.74&58.30&\underline{\textit{\textbf{48.40}}}& \underline{\textit{58.33}}&\underline{\textit{30.23}}&36.13 & 82.95 & 89.74 \\\hdashline

\multirow{2}{*}{IMDB}  & Synthetic   &48.82&61.77&43.09&58.90&45.28&55.59&29.91&35.81 & 79.86 & 86.79\\
                       & + SQuAD & 49.56	&63.10&43.40&59.37&46.68&57.27&30.06&36.05 & 83.33 & 89.92 \\
\hline
\multirow{2}{*}{All 4 data sources} &  Synthetic & 53.28 & 66.32 & 43.64 & 60.43 & 47.41 & 57.88 & 29.91 & 36.37 & 82.71 & 89.06 \\ 
 & + SQuAD & \textbf{53.30} & \textbf{66.73} & \textbf{44.23} & \textbf{60.79}  & 47.01 & \textbf{58.35} & \textbf{30.36} & \textbf{36.50} & \textbf{84.57} & \textbf{90.90} \\ 
\hline
\end{tabular}
}
\caption{{\small Cross domain experiments using QAGen2S as the generative model. \underline{Underlined} cells indicate best EM/F1 value for each of the target domain dev sets (column-wise) and individual target domain corpus.}}\label{tab:crossda}
\end{table*}

\subsection{Domain Adaptation Results}\label{sec:uda}
Tab. \ref{tab:uda} shows the results of domain adaptation experiments. Each experiment was performed by training the RC model on the synthetic data generated on the target domain corpus. We refer to the dataset to which the downstream model is being adapted as the target domain. Source domain indicates the supervised training dataset (SQuAD).

We also performed experiments by using both Synthetic + SQuAD1.1 data. Our QAGen and QAGen2S models outperform by wide margins the baseline models trained on SQuAD 1.1 only, as well as unsupersived domain adaptation approaches (UDA) suggested by \citet{nishida2019uda} and \citet{vaeQA}. Additionally, QAGen and QAGen2S significantly outperforms QGen, our implementation of the three-stage pipeline of \citet{puri2020nvidia}.

Even though our SQuAD 1.1 baselines are generally higher than both \citet{nishida2019uda} and \citet{vaeQA}, our best model achieves more point-wise improvements in all of the target domain datasets, except with BioASQ, where \citet{nishida2019uda} observe 4.3 points in EM versus 4 points with ours, and 4.2 points in F1 versus 2.2 with ours.

Comparing LM and round-trip filtering when applied to the best performing model, QAGen2S, we can observe that the LM filtering approach (Sec. \ref{sec:filtering}) is more effective than round-trip filtering in BioASQ and DuoRC target domains. It barely underperforms ($\sim$ 1 point) in F1 and EM in the other two domains. This demonstrates the efficacy of the suggested filtering approach, which also simplifies the question-answer generation pipeline.

The highest (EM/F1) domain adaptation gains seen with BioASQ (4/2.2) and DuoRC (1.2/1.1) are smaller than those with Natural Questions (8.5/7.5) and NewsQA (5.5/4.5). We postulate this is due to two reasons: Firstly, both BioASQ and DuoRC domains are more dissimilar to the source domain, SQuAD, compared to NewsQA and Natural Questions; Secondly, BioASQ and DuoRC are more difficult  datasets. Comparing our results with supervised target domain training of DuoRC, we observe that with using only synthetic data  outperforms the DuoRC training set, which consists of 39144 pairs.
While our domain adaptation methods show substantial gains with NewsQA and Natural Questions domain, there is still room for improvements to match the performance of supervised target domain training (last row in Tab. \ref{tab:uda}). 

While results in Tab. \ref{tab:uda} suggest that generating synthetic QA data from target domain text leads to significant gains on the target domain dev set, 
one can argue whether it is essential to generate synthetic data from the corpus matching the target dev set's domain to achieve good performance.
Hence, we performed cross-domain experiments to check this argument.
Tab. \ref{tab:crossda} shows the performance on every target domain dev set of RC models fine-tuned on synthetic data of different target domain corpora.
We can see that diagonal elements, 
which have same domain of dev set and target corpus, 
show either the best performance (\underline{underlined} results) or are within a narrow margin of top EM/F1 scores. Therefore, the most effective strategy is achieved when the passages used in the generation of synthetic samples are from the same domain as the target, which is expected in a domain adaptation method.
Additionally, 
we trained an RC model with the synthetic data from all the four domains (last two rows in Tab. \ref{tab:crossda}).
This produced our best F1 results for all datasets,
indicating that mixing synthetic data from different domains is beneficial for the QA task. Tab. \ref{tab:crossda} also shows EM/F1 scores of the cross-domain RC models on SQuAD 1.1 dev set.
We can see that using synthetic data from any of the four domains significantly improved the performance for SQuAD.
In particular, when training the RC model with data from all domains + SQuAD training data (last row), there is a large gain in both EM (3.8) and F1 (2.7).



\begin{table*}[ht]
\centering
\footnotesize
\scalebox{0.92}{
\begin{tabular}{l|ccccccccc}
\hline
\multirow{3}{*}{\textbf{Model}} &
\multicolumn{2}{c}{\textbf{Beam Search}} &
\multicolumn{2}{c}{\textbf{Topk+Nucleus}} & \multicolumn{2}{c}{\textbf{Topk+Nucleus}} & \multicolumn{2}{c}{\textbf{Topk+Nucleus}} \\ 
& \multicolumn{2}{c}{\textbf{N=5}} & \multicolumn{2}{c}{\textbf{N=5}} & \multicolumn{2}{c}{\textbf{N=10}} & \multicolumn{2}{c}{\textbf{N=20}} \\ 

 & EM & F1 & EM & F1 & EM & F1 & EM & F1\\
\hline
Synthetic    & 49.73 & 63.19 & 52.20 & 66.19 & 52.64 & 65.56 & 51.08 & 63.50 \\
Synthetic + SQuAD & 49.95 & 64.08& 49.68 & 64.47 & 52.03 & 65.70 & 51.87 & 64.82 \\
\hline
\end{tabular}
}
\caption{{\small Beam search vs. Topk+Nucleus sampling with various sample sizes per passage. NQ is used as target domain and QAGen2S with LM filtering is used as generator. For N $>$ 5, top 5 samples per passage were selected according to LM scores.}}\label{tab:sampling_options}
\end{table*}

\subsection{Comparison of AQGen, QAGen and QAGen2S models} 
Comparing our proposed LM filtering-based models in Tab. \ref{tab:uda},
we propose the following explanations:
(1) QAGen2S and QAGen outperform AQGen because generating answers conditioned on the question results in better spans, which is crucial in the training of the downstream RC task. Generated answer spans not conditioned on questions could include spurious tokens, or be a partial span.
(2) QAGen2S outperforms QAGen because including the generated question in the bidirectional encoder allows cross attention between the passage and generated question, which results in even more accurate answer generation.
Comparing the performance when only synthetic question-answer pairs are employed versus adding SQuAD training pairs, we can observe that the addition of labeled data results in marginal gains. This becomes even more evident for the best performing data generators. In fact, in some cases, adding SQuAD data degrades EM, such as QAGen2S + LM filtering with Natural Questions and NewsQA.

\subsection{Ablation Studies}
\subsubsection*{Sampling Design Choices}
Tab. \ref{tab:sampling_options} shows a comparison between beam search and  Topk+Nucleus sampling with different  number  of  samples (5, 10, and  20).
The results indicate that beam search underperforms Topk+Nucleus. 
We attribute this to the lack of diversity in the generated samples using beam search.
We observed that beam search tends to select fewer distinct spans, compared to Topk+Nucleus, and generates minor variations of the same question. Appendix \ref{ap:sec:illust_answer_lm_score} examines this issue.

 When training the RC model we only used the top 5 samples based on LM score per each passage.
We can observe that sampling 10 pairs per document leads to the best EM/F1 on the target domain.
By sampling many QA pairs per passage, we increase the chance of generating good samples.
However, if we sample too many qa pairs the top ranked ones might be too similar.
Therefore, we used sample size of 10 in this work since a higher sample size incurs higher computation cost while not showing improvements.


\subsubsection*{LM Filtering}
We argue that using LM filtering, as discussed in section \ref{sec:filtering}, results in improvements in the target domain downstream RC models by enhancing the quality of the generated $(q,a)$ pairs. Results in Tab. \ref{tab:lmfiltering_main} indicate that in the majority of the experiments using LM filtering leads to improved F1/EM metrics. AQGen benefits the most from LM filtering as it generates data with lower quality than the other two models.
Tables \ref{tab:samples_lm_scores} and \ref{tab:table_samples} in the Appendix show examples of QA pairs and their LM scores.

\begin{figure}[htbp]
 \centering
  \includegraphics[width=0.4\textwidth]{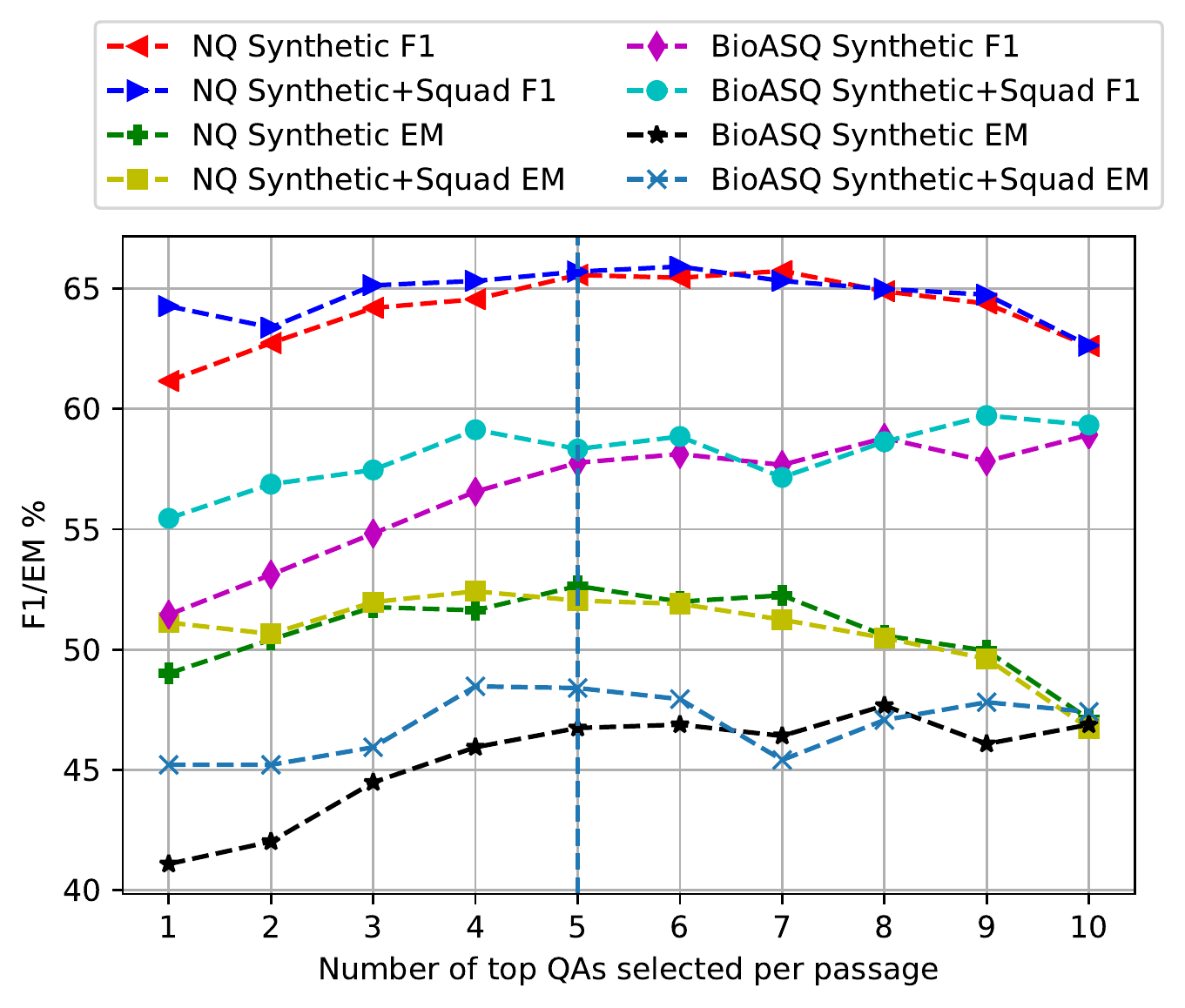}
 \caption{\small Effect of number of QAs selected per passage in LM filtering. QAGen2S model is used for generation. The likelihood score of the generated answer is used to sort the generated question answer pairs decreasingly.}
 \label{figs:local_filtering}
 \end{figure}

Fig. \ref{figs:local_filtering} shows experimental results when varying the number of $(q,a)$ pairs selected from the 10 pairs sampled per each passage.
We chose the value of 5 as this configuration outperforms other values overall. A high value is more likely to allow undesired pairs, while a low value might discard plenty of high quality samples.


\begin{table}[ht]
\centering
\footnotesize\addtolength{\tabcolsep}{-2pt}
\scalebox{0.92}{
\begin{tabular}{lc|cccc}
\hline
\multirow{2}{*}{\textbf{Model}} & \textbf{FT} & \multicolumn{2}{c}{\textbf{NQ}} &  \multicolumn{2}{c}{\textbf{BioASQ}}  \\ 
& \textbf{Data} & EM & F1 & EM & F1 \\
\hline
\multirow{2}{*}{AQGen  w/o filter.}  & Synth.   &46.93&60.71 &41.49&53.59\\ 
                       & + SQ &46.84&61.00 &41.36&53.84 \\ 
                       \hdashline
\multirow{2}{*}{AQGen  + LM filter.}  & Synth.   & 47.80 & 61.29 & 39.49 & 52.11   \\ 
                       & + SQ & \textbf{49.04} & \textbf{62.56} & \textbf{42.89} & \textbf{54.90}  \\ 
                      \hline
                      
\multirow{2}{*}{QAGen w/o filter.}  & Synth.  &50.67&64.04&43.15&53.20
 \\  
                       & + SQ &\textbf{51.35}&\textbf{64.99}& 45.21&54.94\\ \hdashline
                                             
\multirow{2}{*}{QAGen  + LM filter.}  & Synth.   & 49.81 &63.36 & 42.49 & 51.95 \\ 
                       & + SQ &  50.01 & 63.10 & \textbf{45.74} &\textbf{55.06} \\  
\hline
                       
\multirow{2}{*}{QAGen2S w/o filter.}  & Synth. &47.12&62.61&46.88&58.92
  \\
                       & + SQ  &46.73&62.63&47.41&\textbf{59.33}
 \\
\hdashline
\multirow{2}{*}{QAGen2S + LM filter.}  & Synth.   & \textbf{52.64} &65.56& \textbf{48.40}	& 58.33 \\
                       & + SQ &52.03 & \textbf{65.70} & 46.74	& 57.76 \\
\hline
\end{tabular}
}
\caption{{\small Comparison of using LM filtering versus no filtering. \textbf{Bold} values indicate best performance on each target domain for each model (per rows separated by sold lines).}}\label{tab:lmfiltering_main}
\end{table}

\subsubsection*{Correlation between LM and F1 Scores}
\begin{figure}[h]
 \centering
  \includegraphics[width=0.5\textwidth]{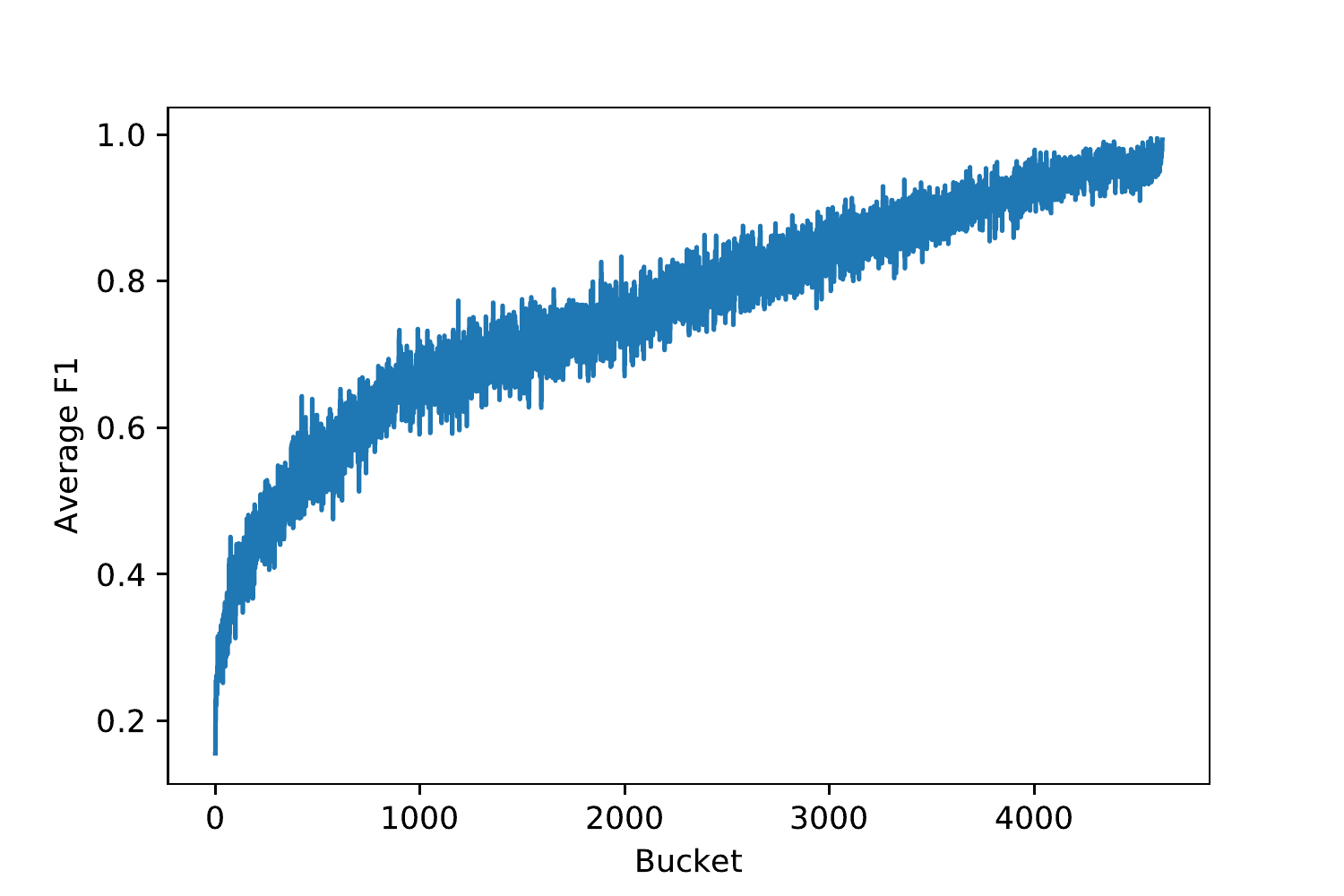}
 \caption{\small Average F1 score of sorted items based on LM scores. Samples were generated using QAGen2S on Natural Questions passages.}
 \label{figs:lmvsf1}
 \end{figure}

In this work, we proposed using the LM score of the generated samples as a surrogate to round-trip filtering. We postulate that the LM score correlates with the F1 score used in round-trip filtering. To more thoroughly examine this, we devised an experiment where we sorted the generated samples by their answer LM scores, divided them into contiguous buckets each with 200 samples, and calculated the average F1 score of the samples in each bucket. Fig. \ref{figs:lmvsf1} shows the results of this experiment. As we can see, there exists a strong correlation between the two scores.

While the correlation looks promising, a challenge with using the LM score is that it is relatively noisy. For example, to use the LM score to get only samples whose F1 scores are 1, a very high threshold needs to be set, forcing the vast majority of samples to be dropped. Future work can explore how to reduce this noise.

\subsubsection*{Impact of Synthetic Dataset Size}
\begin{figure}[htbp]
 \centering
  \includegraphics[width=0.4\textwidth]{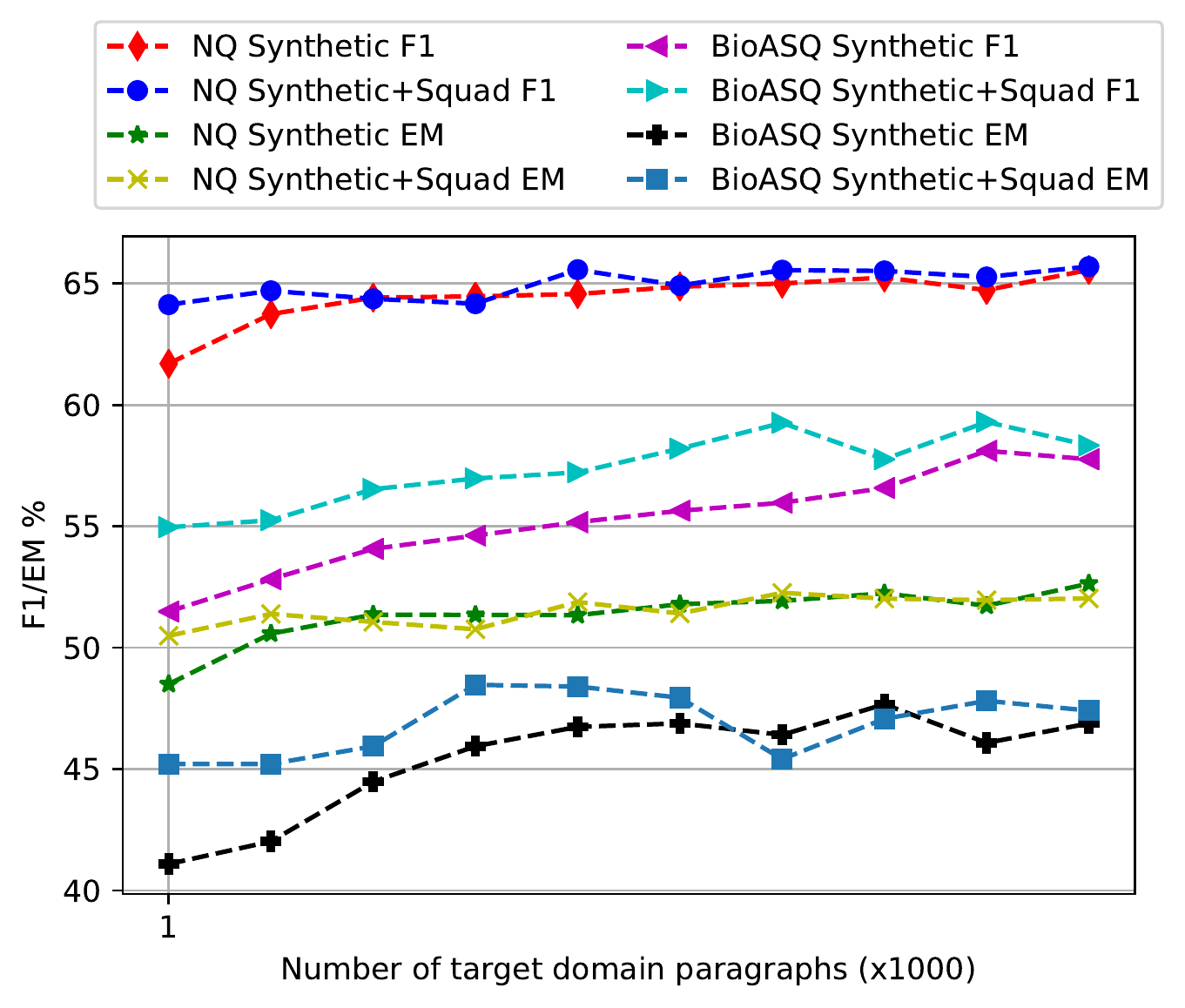}
 \caption{\small The effect of number of target domain passages on the RC task with synthetically generated QA pairs. QAGen2S is employed to generate questions on NQ and PubMed.}
 \label{figs:parasvsf1}
 \end{figure}

In Fig. \ref{figs:parasvsf1}, we present plots that correlate synthetic dataset size (in \# of passages) and RC model performance (EM/F1). We can see that with increasing the number of generated $(q,a)$ pairs (5 pairs per passage), RC model performance improves. Such correlation is more evident when not using the SQuAD training data. This is expected as with added supervised training samples, there would be less need for a large number of synthetic samples.

\subsection{Experiments with Large QA Models}
The downstream RC models presented in previous sections were based on fine-tuning \texttt{BERT-base} model, which has 110 million parameters. In this section, we assess the efficacy of our proposed domain adaptation approach on a higher capacity transformer as the RC model. For these experiments, 
we chose pretrained \texttt{RoBERTa-large} \cite{roberta} model from transformers library \cite{HuggingFacesTS}, which has 355 million parameters.
Tab. \ref{tab:roberta} displays the domain adaptation results on the NQ domain using QAGen2S generated samples. It also includes performance on the source domain dev set. 
Although the SQuAD 1.1 baselines (first row), is significantly higher than those with \texttt{BERT-base} in Tab. \ref{tab:uda}, EM/F1 gains of 5.8/3.4 are achieved on the target domain. 1/0.5 gains in EM/F1 are observed in SQuAD 1.1 dev set. These results demonstrate that our proposed end-to-end synthetic data generation approach is capable of achieving substantial gains even on state-of-the-art RC baselines such as \texttt{RoBERTa-large}.

\begin{table}[ht]
\centering
\footnotesize\addtolength{\tabcolsep}{-2pt}
\scalebox{0.92}{
\begin{tabular}{lc|cccccccc}
\hline
\multirow{2}{*}{\textbf{Model}} & \textbf{FT} & \multicolumn{2}{c}{\textbf{SQuAD 1.1}} &
\multicolumn{2}{c}{\textbf{NQ}} \\ 
& \textbf{Data} & EM & F1 & EM & F1 \\
\hline
SQuAD1.1 (SQ) & SQ & 86.43 & 93.18 & 50.57 & 67.09
\\
\hdashline
\multirow{2}{*}{QAGen2S w/o filter.}  & Synth. & 85.39	
 &  92.15 &51.20 & 	67.25 
  \\
                       & + SQ  & 86.23 & 93.19 & 50.73 & 67.07
 \\
\hdashline
\multirow{2}{*}{QAGen2S + LM filter.}  & Synth. & 85.77 & 92.07

 & 55.06 & 68.83  \\
                       & + SQ & 86.75 & 93.50 & 55.73 & 70.04\\
                       \hdashline

\multirow{2}{*}{QAGen2S + RT filter.}  & Synth.   & 85.80 &  92.15 & 56.46 & 70.39 \\
                       & + SQ & \textbf{87.46} & \textbf{93.67} & \textbf{56.35} & \textbf{70.47}\\
\hline
\end{tabular}
}
\caption{{\small Source and target domain performance with RoBERTa-large as downstream RC model.}}\label{tab:roberta}
\end{table}

%% file: conclusion.tex
\vspace{-.5cm}
\section{Conclusions}
We presented a novel end-to-end approach to generate question-answer pairs by using a single transformer-based model. Our experiments showed that by proper decoding, significant improvements in domain adaptation of RC models can be achieved.  We concluded that using LM filtering improves the quality of synthetic question-answer pairs; however, there is still a gap with round-trip filtering with some of the target domains. Improving LM-score-based filtering is a future direction of our work.

While we were able to generate diverse, high quality and challenging synthetic samples on the target domains, the types of the questions produced still were limited to those of SQuAD, since the generative models were trained on SQuAD. It would be interesting to explore how one can adapt the generative models to the type of target domain questions.

%% file: appendix.tex
\appendix 
\section{Additional Details Regarding the Datasets Used}
\textbf{SQuAD 1.1} dataset  is used to train the generative models as well as in-domain supervised data for the downstream RC task. We use the default train and dev splits, which contain 87,599 and 10,570 $(q,a)$ pairs, respectively. SQuAD 1.1 questions exhibit high lexical overlap with answers, since annotators were presented with passages and extracted answers when creating questions. \\
\textbf{Natural Questions} dataset consists of Google search questions, the Wikipedia pages from top 5 search results, and the corresponding annotated answers. 
This dataset and SQuAD are both derived from Wikipedia pages, however, questions from Natural Questions have considerably less ngram overlap with annotated answers compared to those from SQuAD.
Also different from SQuAD, Natural Questions dataset contains passages with HTML tables and tags. We use MRQA Shared Task preprocessed training and dev sets, which consist of 104,071 and 12,836 $(q,a)$ pairs, respectively. We utilize training set passages as the target domain (unlabeled) corpus, while preforming evaluations on the  dev set.\\
\textbf{NewsQA} consists of question and answer pairs from CNN news articles. We use the dev set from the MRQA Shared Task, which removes unanswerable questions and those without annotator agreement. We believe this version better suits our work, as we focus only on generation of answerable questions. The train and dev sets consists of 74,160 and 4,212 samples, respectively. Passages from CNN/Daily Mail corpus are used as target domain passages.\\
\textbf{BioASQ} challenge is a competition on semantic indexing and question answering tasks based on annotated PubMed documents. As with the previous dataset, we employ MRQA shared task version of BioASQ, which consists of a dev set with 1,504 pairs. We collected PubMed abstracts to use as target domain passages. Being from Biomedical domain, BioASQ makes a clear domain shift from other datasets. \\
\textbf{DuoRC} contains question answer pairs from movie plots which are extracted from both Wikipedia and IMDB. This dataset has been developed to have question and answer pairs with minimal lexical overlap, which makes it more challenging.
ParaphraseRC task of DuoRC dataset is used in our evaluations. Training and dev sets include 39,144 and 13,111 pairs, respectively. We crawled IMDB movie plots to use as the target domain unlabeled corpus.
The dataset has been developed by selecting the same movie plot from both sources, and generating question from one source and selecting the answer from the other. This approach has resulted in question and answer pairs with minimal lexical overlap. 

All of the validation sets of the aforementioned out-of-domain tasks are identical to those used by Nishida et al., except DuoRC, where we use MRQA shared task formatted DuoRC dev set.
\section{Additional Ablation Studies}
\begin{table*}[h]
\centering
\footnotesize
\scalebox{0.92}{
\begin{tabular}{lc|cccccccc}
\hline
\multirow{2}{*}{\textbf{Model}} & \textbf{fine-tune} & \multicolumn{2}{c}{\textbf{None}} & \multicolumn{2}{c}{\textbf{LM}} & \multicolumn{2}{c}{\textbf{Rountrip}} \\ 
& \textbf{Data} & EM & F1 & EM & F1 & EM & F1\\
\hline
\multirow{2}{*}{QGen}  & Synthetic& 70.31 & 80.34 & -- & -- & 77.11 & 84.81\\
                       & + SQuAD &  81.50 & 89.01 & -- & -- & \textbf{82.94} & \textbf{89.68}\\
\hline
\multirow{2}{*}{AQGen}  & Synthetic   &74.58 &	84.14 & 74.34 &	83.55 & 78.51 &	86.21 \\ 
                       & + SQuAD & 82.10 &	89.47 & 82.15&	89.31 & \textbf{82.88}&	\textbf{89.78} \\ \hline
\multirow{2}{*}{QAGen}  & Synthetic   & 79.65&	87.14 &78.40 &	85.98 & 78.51&	86.21\\ 
                       & + SQuAD & \textbf{83.07}&\textbf{90.00} & 82.53&89.51 & 83.03&89.74 \\ \hline
\multirow{2}{*}{QAGen2S}  & Synthetic   &81.25&88.20 & 79.86&86.79 & 80.61&87.36
 \\
& + SQuAD  & \textbf{83.87}&\textbf{90.40}& 83.33 &	89.92 & 83.29 & 89.84 \\
\hline
\end{tabular}
}
\caption{{\small Performance on SQuAD 1.1 development set when training with LM-filtered synthetically generated question-answer pairs on IMDB corpus. \textbf{Bold} values indicate best performance per each model (row-wise). Our baseline EM and F1 numbers (on SQuAD 1.1 training set) are 80.78 and 88.20, respectively.
}}\label{tab:squad}
\end{table*}
\subsection{Performance on SQuAD 1.1 with Different Filtering Approaches}

While the performance of the RC models on the target domains is important, weak performance on the source domain could inhibit the use of our proposed methods in applications that require strong performance in both source and target domains. 
Tab. \ref{tab:squad} shows EM/F1 scores of the \texttt{bert-base-uncased} RC models trained with synthetic data generated from the IMDB corpus on SQuAD 1.1 dev set. We can observe that adding synthetic samples to the SQuAD training set always improves the performance on the dev set compared to using the SQuAD training set only. In fact, with QAGen2S,  impressive  3.1(EM)/2.2(F1) gains are achieved. Synthetic only samples from the same model outperform the SQuAD baseline.
Similar to previous domain adaptation results, we observe that QAGen2S outperforms QAGen, and QAGen exceeds AQGen. 
\subsection{Comparison of Using Filtering vs. No Filtering}
Tab. \ref{tab:lmfiltering} presents comprehensive results of using LM filtering over all the of the target domains. We can observe that the arguments made in Sec. 4.7 hold for NewsQA and DuoRC as well. 
\begin{table*}[h]
\centering
\footnotesize
\scalebox{0.92}{
\begin{tabular}{lc|ccccccccc}
\hline
\multirow{2}{*}{\textbf{Model}} & \textbf{fine-tune} & \multicolumn{2}{c}{\textbf{NQ}} & \multicolumn{2}{c}{\textbf{NewsQA}} & \multicolumn{2}{c}{\textbf{BioASQ}} &
\multicolumn{2}{c}{\textbf{DuoRC}} & \textbf{Synthetic} \\ 
& \textbf{Data} & EM & F1 & EM & F1 & EM & F1 & EM & F1 & \textbf{\#}\\
\hline
\multirow{2}{*}{AQGen  w/o filtering}  & Synthetic   &46.93&60.71&36.21 & 53.83 &41.49&53.59&26.94&33.46 & \multirow{2}{*}{860k}\\ 
                       & + SQuAD &46.84&61.00 &36.99 & 54.47 &41.36&53.84&26.87&33.43 \\ 
                       \hdashline
\multirow{2}{*}{AQGen  + LM filtering}  & Synthetic   & 47.80 & 61.29 & 38.56	& 55.42 & 39.49 & 52.11 & 27.09	& 33.47 & \multirow{2}{*}{490k}  \\ 
                       & + SQuAD & \textbf{49.04} & \textbf{62.56} & \textbf{39.62} & \textbf{56.89} & \textbf{42.89} & \textbf{54.90} & \textbf{27.88} & \textbf{34.40}  \\ 
                      \hline
                      
\multirow{2}{*}{QAGen w/o filtering}  & Synthetic  &50.67&64.04&43.07 & \textbf{59.53}&43.15&53.20&29.68&35.78  & \multirow{2}{*}{890k}
 \\  
                       & + SQuAD &\textbf{51.35}&\textbf{64.99}& 42.64 & 59.4 &45.21&54.94&29.87&\textbf{35.87}\\ \hdashline
                                             
\multirow{2}{*}{QAGen  + LM filtering}  & Synthetic   & 49.81 &63.36 & 43.1	& 57.94 & 42.49 & 51.95& 29.46 &35.25  & \multirow{2}{*}{500k} \\ 
                       & + SQuAD &  50.01 & 63.10 & \textbf{44.06} &	59.20 & \textbf{45.74} &\textbf{55.06} & \textbf{29.91}	& 35.82 \\  
\hline
                       
\multirow{2}{*}{QAGen2S w/o filtering}  & Synthetic &47.12&62.61&43.38 & 60.1&46.88&58.92&30.04&\textbf{36.58}  & \multirow{2}{*}{890k}
  \\
                       & + SQuAD  &46.73&62.63&43.87 & \textbf{60.51}&47.41&\textbf{59.33}&30.00&36.49
 \\
\hdashline
\multirow{2}{*}{QAGen2S + LM filtering}  & Synthetic   & \textbf{52.64} &65.56& \textbf{43.99} &	59.94& \textbf{48.40}	& 58.33 & 29.91	& 35.81  & \multirow{2}{*}{480k} \\
                       & + SQuAD &52.03 & \textbf{65.70} & 43.57 &59.8 & 46.74	& 57.76 & \textbf{30.06} & 36.05 \\
\hline
\end{tabular}
}
\caption{{\small Comparison of using LM filtering versus no filtering. \textbf{Bold} values indicate best performance on each target domain for each model (per rows separated by sold lines).}}\label{tab:lmfiltering}
\end{table*}

\subsection{Impact of Language Model Score Pooling}
\label{ap:sec:lm_score_pooling}
To aggregate the LM scores of a given question-answer pair, one can use either sum or average of the token scores, as defined in Sec. \ref{sec:filtering}. We experimented with both options and summarized the results in Tab. \ref{tab:lmpooling} for QAGen and AQGen models. We can observe that using summation generally outperforms averaging. We speculate this is because average pooling encourages longer question-answer pairs, which are more likely to consist of incorrect samples. By using summation, shorter question-answer pairs would be more likely to be selected during LM filtering.
\begin{table*}[!ht]
\centering
\footnotesize
\scalebox{0.92}{
\begin{tabular}{lcc|cccccccc}
\hline

\multirow{2}{*}{\textbf{Model}} & \textbf{LM} & \textbf{fine-tune} & \multicolumn{2}{c}{\textbf{NQ}} & \multicolumn{2}{c}{\textbf{NewsQA}} & \multicolumn{2}{c}{\textbf{BioASQ}} &
\multicolumn{2}{c}{\textbf{DuoRC}}  \\ 
 & \textbf{Pooling} & \textbf{Data} & EM & F1 & EM & F1 & EM & F1 & EM & F1 \\
\hline
\multirow{4}{*}{AQGen} & \multirow{2}{*}{Sum} & Synthetic   & 47.80 & 61.29 & 38.56	& 55.42 & 39.49 & 52.11 & 27.09	& 33.47   \\ 
                       & & + SQuAD & \textbf{49.04} & \textbf{62.56} & \textbf{39.62} & \textbf{56.89} & \textbf{42.89} & \textbf{54.90} & \textbf{27.88} & \textbf{34.40}  \\ 
                       \cdashline{2-11}
  & \multirow{2}{*}{Avg} & Synthetic   &47.73 &	61.98   & 34.19	& 52.05 & 39.03	& 51.52 & 26.31 &	32.84   \\ 
                       & & + SQuAD &  45.03	& 59.87 & 35.21 & 53.08  & 40.82 & 53.74 & 26.7 & 33.26 \\ 
                      \hline
                      
\multirow{4}{*}{QAGen} & \multirow{2}{*}{Sum} & Synthetic   & 49.81 &63.36 & 43.1	& 57.94 & 42.49 & 51.95& 29.46 &35.25 \\ 
                       &  & + SQuAD &  50.01 & 63.10 & \textbf{44.06} &	\textbf{59.20} & \textbf{45.74} &\textbf{55.06} & \textbf{29.91}	& \textbf{35.82} \\  
                       \cdashline{2-11}
  & \multirow{2}{*}{Avg} & Synthetic   &  \textbf{50.3} &  \textbf{63.93}  & 	43.14 & 58.82 &  41.82 &	52.22 & 28.5 & 34.51   \\ 
                       & & + SQuAD &  50.18	& 63.71 & 42.76	& 58.65  & 42.15 & 52.21 &  29.01	& 35.05  \\ 
                      \hline
\end{tabular}
}
\caption{{\small Comparison of using average versus summation of LM scores when doing LM filtering. \textbf{Bold} values indicate the best performance on each target domain for each model (per rows separated by solid lines).}}\label{tab:lmpooling}
\end{table*}

\begin{table*}[h]
\centering
\scriptsize
\begin{tabular}{lll}
 \hline
\multicolumn{3}{l}{}
\begin{minipage}{15 cm}
\textit{\textbf{Passage:}}
\\
 <P> The United States is estimated to have a population of \textbf{327,589,916} as of April 23 , 2018 , making it the \textbf{third} most populous country in the world . It is very urbanized , with \textbf{81 $\%$} residing in \textbf{cities and suburbs} as of 2014 ( the worldwide urban rate is 54 $\%$ ) . \textbf{California and Texas} are the most populous states , as the mean center of U.S. population has consistently shifted westward and southward . \textbf{New York City} is the most populous city in the United States . </P>
  \end{minipage} \hfill
  \\
 \hdashline
 \textbf{Topk+Nucleus} \\
 Q: \textit{As of April 23, 2018, what is the estimated population of the US?} & A: \textit{327,589,916} & LM score: \textit{-0.00577}\\ 
 Q: \textit{How many people lived in the US in April of 2018?} & A:  \textit{327,589,916}& LM score: \textit{-0.00707} \\ 
 Q: \textit{What is the population of the United States?} & A:  \textit{327,589,916}& LM score: \textit{-0.01358} \\ 
 Q: \textit{What is the most populous city in the United States?} & A:  \textit{New York City}& LM score: \textit{-0.04131} \\
 Q: \textit{Where do 81 percent of Americans live?} & A:  \textit{cities and suburbs}& LM score: \textit{-0.05360} \\
 Q: \textit{Where does the United States rank among most populous countries on the planet?} & A:  \textit{third}& LM score: \textit{-0.07449} \\
 Q: \textit{How much of the US's population is concentrated in the metropolitan areas of the country?} & A:  \textit{81 $\%$}& LM score: \textit{-0.09509} \\
 Q: \textit{How much of the US population is urbanized?} & A:  \textit{81 $\%$}& LM score: \textit{-0.1375} \\
 Q: \textit{What two cities have the highest populations in America?} & A:  \textit{California and Texas}& LM score: \textit{-0.18128} \\
 Q: \textit{What country is considered the most populous?} & A:  \textit{third}& LM score: \textit{-1.85929} \\  \hdashline
  \textbf{Beam Search} \\
 Q: \textit{What is the population of the United States as of April 23, 2018?} & A: \textit{327,589,916} & LM score: \textit{-0.00492}\\ 
 Q: \textit{As of April 23, 2018, what was the population of the United States?} & A: \textit{327,589,916} & LM score: \textit{-0.00529}\\ 
 Q: \textit{As of April 23, 2018, how many people live in the United States?} & A: \textit{327,589,916} & LM score: \textit{-0.00618}\\ 
 Q: \textit{How many people live in the United States?} & A: \textit{327,589,916} & LM score: \textit{-0.0132}\\ 
 Q: \textit{What is the population of the United States?} & A: \textit{327,589,916} & LM score: \textit{-0.0135}\\ 
 \hline
\multicolumn{2}{l}{}
\end{tabular}
\caption{\small Samples of generated question-answers pairs using QAGen2S model from Natural Questions passages with their LM scores. Sum of answer likelihood scores is used to sort the pairs decreasingly. The generated answers are shown in \textbf{bold}. Samples shown  from Beam Search with beam size of 5, and Topk+Nucleus with sample size of 10.}
\label{tab:samples_lm_scores}
\end{table*}

\section{Examples of Generated Samples}
\subsection{Illustration of Answer LM Score}
\label{ap:sec:illust_answer_lm_score}
Tab. \ref{tab:samples_lm_scores} presents unfiltered question-answer pairs and associated answer LM scores generated from a randomly selected Natural Questions corpus using the QAGen2S model. As can be seen from Topk+Nucleus decoded samples, the last two generated samples are incorrect and would be filtered out using the LM filtering approach that is used in this work. The last sample, which consists of an answer that is entirely irrelevant to its question, has a considerably lower answer LM score than the rest of the samples.

With beam search, due to the high number of repetitions, the scores are close. While beam search generates samples with high likelihood, due to the lack of diversity, as evident here, the performance of the trained RC models on such synthetic samples underperforms those of Topk+Nucleus.
\begin{table*}[h]
\centering
\scriptsize
\begin{tabular}{ll}
 \hline
\multicolumn{2}{l}{}
\begin{minipage}{15 cm}
\textit{\textbf{Passage:}}
(CNN) -- Fifteen people have now died after consuming \textbf{cantaloupe} contaminated with the listeria monocytogenes bacteria, the Centers for Disease Control and Prevention said \textbf{Friday}. At least \textbf{84} people in 19 states have become ill with the bacteria, the agency said. And the \textbf{number of illnesses} could still grow, added the CDC, citing \textbf{reporting lags and how the disease can develop slowly in some people}. On Tuesday, the CDC was reporting 13 deaths and 72 illnesses in what was already then the deadliest food-borne illness outbreak in the United States since \textbf{1998}. Five people have died in New Mexico from eating the tainted cantaloupes, the CDC said. \textbf{Three} people died in Colorado, \textbf{two} in Texas and one each in Kansas, Maryland, Missouri, Nebraska and Oklahoma. Illnesses have also been reported in Alabama, Arkansas, California, Illinois, Indiana, Montana, North Dakota, Virginia, West Virginia, Wisconsin and Wyoming. What you need to know about Listeria. \textbf{Most of those who fell ill are more than 60 years old}, the CDC said. Doctors also are closely monitoring the pregnancies of two women who ate contaminated cantaloupe, with the agency noting that \textbf{listeriosis} can cause miscarriages and stillbirths. Older adults and people with compromised immune systems are also especially susceptible. Public health investigators have traced the source of the bacteria to a farm in \textbf{Granada, Colorado}. Food Poisoning 101. The grower, \textbf{Jensen Farms}, issued a recall for its Rocky Ford-brand cantaloupes on September 14. By now, the cantaloupes should all be off store shelves, the CDC said. The agency warned that people should not eat Rocky Ford cantaloupes, even if they have eaten part of one and have not yet fallen ill. It also said that consumers \textbf{should be wary of eating any cantaloupes if they don't know where they came from}. How to keep your food safe.
 \end{minipage} \hfill \\
 \hdashline
  \textbf{\textit{AQGen}} :\\
 Q: \textit{What can cause miscarriages?} & A: \textit{listeriosis} \\ 
 Q: \textit{Which state has had the most deaths?} & A:  \textit{Colorado} \\ 
 Q: \textit{Where is the farm where the bacteria came from?} & A:  \textit{Colorado} \\ 
 Q: \textit{How many people have died from eating listeria from cantaloupe?} & A:  \textit{14} \\
 Q: \textit{Where has the worst case happened?} & A:  \textit{Colorado} \\
 Q: \textit{Where were the listeria monocytogenes bacteria come from?} & A:  \textit{Granada} \\
  \hdashline
 \textbf{\textit{QAGen}} :\\
 Q: \textit{What year was the deadliest food-borne illness outbreak in the United States since?} & A: \textit{1998} \\ 
 Q: \textit{How old were most of the victims of the outbreak?} & A:  \textit{more than 60 years old} \\ 
 Q: \textit{How old were most of the people who died from the listeria infection?} & A:  \textit{more than 60 years old} \\ 
 Q: \textit{How many people in the US have become seriously ill with Listeria?} & A:  \textit{84} \\
 Q: \textit{How many people in Texas were killed by tainted cantaloupes?} & A:  \textit{two} \\
 Q: \textit{How old were most of the people who died from the listeria infection?} & A:  \textit{more than 60} \\
 Q: \textit{How many people were reported killed in Colorado?} & A:  \textit{Three} \\
 Q: \textit{Where has the food poisoning been traced to?} & A:  \textit{Granada, Colorado} \\
 Q: \textit{Who did the CDC have in custody over the tainted cantaloupes?} & A: \textit{Jensen Farms} \\ 
 Q: \textit{Who released the recall announcement?} & A:  \textit{Jensen Farms} \\ 
 \hdashline
 \textbf{\textit{QAGen2S}} :\\
 Q: \textit{What can cause miscarriages and stillbirths?} & A: \textit{listeriosis} \\ 
 Q: \textit{What type of food was it?} & A:  \textit{cantaloupe} \\ 
 Q: \textit{What was the first year of death from this outbreak?} & A:  \textit{1998} \\ 
 Q: \textit{How does the food-borne illness outbreak effect those over 60?} & A:  \textit{Most of those who fell ill are more than 60 years old} \\
 Q: \textit{When did the CDC start reporting the Listeria monocytogenes bacteria in cantaloupes?} & A:  \textit{Friday} \\
 Q: \textit{How old are most of those in the recent outbreak?} & A:  \textit{more than 60 years old} \\
 Q: \textit{How could the number of sickened listeria possibly grow?} & A:  \textit{reporting lags and how the disease can develop slowly in some people} \\
 Q: \textit{When did the CDC start reporting the Listeria monocytogenes bacteria?} & A:  \textit{Friday} \\
 Q: \textit{What could still grow?} & A:  \textit{number of illnesses} \\
 Q: \textit{How can listeriosis be avoided?} & A:  \textit{should be wary of eating any cantaloupes } \\ 
 &\textit{if they don't know where they came from} \\
 \hline
\multicolumn{2}{l}{}
\end{tabular}
\caption{\small Samples of generated question-answers pairs from randomly selected passage from CNN/Daily Mail corpus. Samples are sorted according to LM scores.}
\label{tab:all_models_samples}
\end{table*}
\subsection{Comparison of Generated Samples by AQGen, QAGen and QAGen2S}
Tab. \ref{tab:all_models_samples} presents unfiltered question-answers pairs generated using each of our proposed models on a randomly selected passage from CNN/Daily Mail corpus. We can observe that generated samples using AQGen have lower quality than the other two models. Also, the selected spans are repetitive. Only 3 out of the 6 properly generated samples are correct question-answer pairs. 
Comparing QAGen and QAGen2S samples, we can observe that QAGen2S generates more diverse and longer answer spans. In this example, we can see that more repeated spans are generated by QAGen than QAGen2S.

While the Topk+Nucleus sampling approach improves the diversity of generated question-answer pairs, we can still see repetitions and incorrect pairs. We believe using the LM score filtering, the vast majority of incorrect pairs are discarded. However, this also means there is room for improving the generative models.

\subsection{Question Answers from Table}
\label{ap:sec:qa_from_table}
\begin{table*}[h]
\centering
\scriptsize
\begin{tabular}{lll}
 \hline
\multicolumn{3}{l}{}
\begin{minipage}{15 cm}
\textit{\textbf{Passage:}}
\\
<Table> <Tr> <Th colspan="2"> Tampa Bay Lightning </Th> </Tr> <Tr> <Td colspan="2"> 2018 -- 19 Tampa Bay Lightning season </Td> </Tr> <Tr> <Td colspan="2"> </Td> </Tr> <Tr> <Th> Conference </Th> <Td> Eastern </Td> </Tr> <Tr> <Th> Division </Th> <Td> Atlantic </Td> </Tr> <Tr> <Th> Founded </Th> <Td> 1992 </Td> </Tr> <Tr> <Th> History </Th> <Td> Tampa Bay Lightning 1992 -- present </Td> </Tr> <Tr> <Th> Home arena </Th> <Td> Amalie Arena </Td> </Tr> <Tr> <Th> City </Th> <Td> Tampa , Florida </Td> </Tr> <Tr> <Td colspan="2"> </Td> </Tr> <Tr> <Th> Colors </Th> <Td> Tampa Bay blue , white </Td> </Tr> <Tr> <Th> Media </Th> <Td> Fox Sports Sun 970 AM </Td> </Tr> <Tr> <Th> Owner ( s ) </Th> <Td> Tampa Bay Sports and Entertainment ( Jeffrey Vinik , chairman ) </Td> </Tr> <Tr> <Th> General manager </Th> <Td> Steve Yzerman </Td> </Tr> <Tr> <Th> Head coach </Th> <Td> Jon Cooper </Td> </Tr> <Tr> <Th> Captain </Th> <Td> Steven Stamkos </Td> </Tr> <Tr> <Th> Minor league affiliates </Th> <Td> Syracuse Crunch ( AHL ) Orlando Solar Bears ( ECHL ) </Td> </Tr> <Tr> <Th> Stanley Cups </Th> <Td> 1 ( 2003 -- 04 ) </Td> </Tr> <Tr> <Th> Conference championships </Th> <Td> 2 ( 2003 -- 04 , 2014 -- 15 ) </Td> </Tr> <Tr> <Th> Presidents ' Trophy </Th> <Td> 0 </Td> </Tr> <Tr> <Th> Division championships </Th> <Td> 3 ( 2002 -- 03 , 2003 -- 04 , 2017 -- 18 ) </Td> </Tr> <Tr> <Th> Official website </Th> <Td> www.nhl.com/lightning </Td> </Tr> </Table>
\\
  \end{minipage} \hfill \\
\multicolumn{3}{l}{}
  \begin{minipage}{15 cm}
\textit{\textbf{Rendered Passage:}}
\\
\includegraphics[width=.6\linewidth]{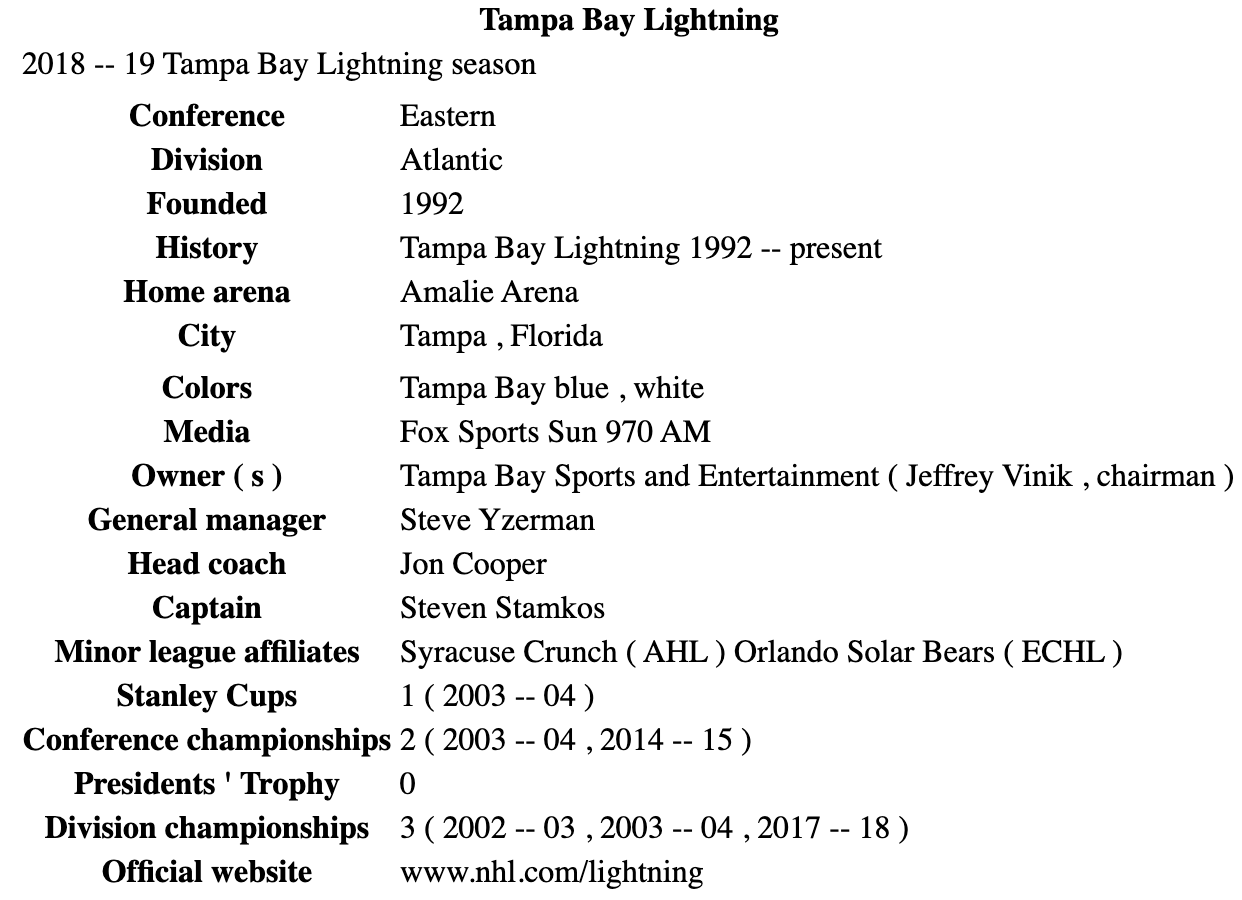}
\\
  \end{minipage} \hfill \\
  \\
 \hdashline
 Q: \textit{What year was the Tampa Bay Lightning established??} & A: \textit{1992} & LM score: \textit{-0.001539}\\ 
 Q: \textit{Who is the head coach of the Tampa Bay Lightning?} & A:  \textit{Jon Cooper}& LM score: \textit{-0.0015659} \\ 
 Q: \textit{Who is the Tampa Bay Lightning general manager?} & A:  \textit{Steve Yzerman}& LM score: \textit{-0.002090} \\ 
 Q: \textit{Who is the Head coach of the Tampa Bay Lightning?} & A:  \textit{Jon Cooper}& LM score: \textit{-0.003044} \\
 Q: \textit{Who is the General Manager of the Tampa Bay Lightning?} & A:  \textit{Steve Yzerman}& LM score: \textit{-0.003877} \\
 Q: \textit{What team is Tampa Bay's home arena?} & A:  \textit{Amalie Arena}& LM score: \textit{-0.00543} \\
 Q: \textit{For whom did Jeffrey Vinik serve as chairman?} & A:  \textit{Tampa Bay Sports and Entertainment}& LM score: \textit{-0.0215854} \\
 Q: \textit{Tampa Bay Sports and Entertainment is owned by what?} & A:  \textit{Jeffrey Vinik}& LM score: \textit{-0.087364} \\

 \hline
\multicolumn{2}{l}{}
\end{tabular}
\caption{\small Generated samples using QAGen2S model from a Natural Questions passage consisting of a table. Sum of answer likelihood scores are chosen to sort the pairs decreasingly.}
\label{tab:table_samples}
\end{table*}
The Natural Questions dataset includes HTML formatted passages. We noticed that some of them are web tables. Tab. \ref{tab:table_samples} illustrates one such example. The content under \textbf{\textit{Passage}} is the input string, as seen by the generative models, and \textbf{\textit{Rendered Passage}} indicates how the table appears in a browser. We experimented with using \textbf{QGen} model on this passage, and noticed that the span detection model was not capable of distinguishing between textual content and HTML tags properly, resulting in selecting spans that included HTML tags. 
However, the samples generated by the joint span and question generation model, QAGen2S in this example, show surprisingly high-quality spans and questions. Only one sample is not correct (\textit{What team is Tampa Bay’s home arena?}). We believe this is because when the span generation is conditioned on the generated question, the likelihood of spans that include spurious tokens, HMTL tags in this example, diminishes sharply.
This opens the door to the possibility of using our proposed models in structured corpora without any extra effort.

\section{Training and Platform Details}
All of the experiments in this work were performed on Amazon EC2 instances. We employed p3.8xlarge, p3.16xlarge, and p3dn.24xlarge GPU instances.
In the training of the generative models, warmup was set to 10$\%$ of total training steps. We used a batch size of 24. Each epoch took 2 to 3 hours on 3 GPUs. We observed that usually, the best model is achieved within the first two epochs. 

The RC models with Synthetic+SQuAD samples were trained by combining synthetic samples and SQuAD training set and randomly shuffling them. Each epoch of training took 2 to 12 hours, depending on the average length of target domain passages on 1 GPU.

All of the hyperparameters of both generative and RC downstream models were fixed. We only performed hyperparameter tuning on those mentioned in the paper.